# Predicting breast tumor proliferation from whole-slide images: the TUPAC16 challenge

Mitko Veta[1], Yujing J. Heng[2], Nikolas Stathonikos[3], Babak Ehteshami Bejnordi[4], Francisco Beca[5], Thomas Wollmann[6], Karl Rohr[6], Manan A. Shah[7], Dayong Wang[2], Mikael Rousson[8], Martin Hedlund[8], David Tellez[4], Francesco Ciompi[4], Erwan Zerhouni[9], David Lanyi[9], Matheus Viana[19], Vassili Kovalev[11], Vitali Liauchuk[11], Hady Ahmady Phoulady[12], Talha Qaiser[13], Simon Graham[13], Nasir Rajpoot[13], Erik Sjöblom[14], Jesper Molin[14], Kyunghyun Paeng[15], Sangheum Hwang[15], Sunggyun Park[15], Zhipeng Jia[16], Eric I-Chao Chang[17], Yan Xu[17,18], Andrew H. Beck[2], Paul J. van Diest[3] and Josien P. W. Pluim[1]

1 Medical Image Analysis Group, Eindhoven University of Technology, Eindhoven, The Netherlands
2 Department of Pathology, Beth Israel Deaconess Medical Center, Harvard Medical School, Boston, MA, USA
3 Department of Pathology, University Medical Center Utrecht, Utrecht, The Netherlands
4 Diagnostic Image Analysis Group, Radboud University Medical Center, Nijmegen, the Netherlands
5 Department of Pathology, Stanford University School of Medicine
6 Biomedical Computer Vision Group, University of Heidelberg, BIOQUANT, IPMB and DKFZ, Heidelberg, Germany
7 The Harker School, San Jose, USA
8 ContextVision AB, Linköping, Sweden
9 Foundations of Cognitive Computing, IBM Research Zürich, Rüschlikon, Switzerland
10 Visual Analytics and Insights, IBM Research Brazil, São Paulo, Brazil
11 Biomedical Image Analysis Department, United Institute of Informatics, Minsk, Belarus
12 Department of Computer Science and Engineering, University of South Florida, Tampa, Florida, USA
13 Department of Computer Science, University of Warwick, Warwick, UK
14 Research, Sectra, Linköping, Sweden
15 Lunit Inc., Seoul, Korea
16 Institute for Interdisciplinary Information Sciences, Tsinghua University, Beijing, China
17 Microsoft Research, Beijing, China
18 Biology and Medicine Engineering, Beihang University, Beijing, China
19 IBM Research Brazil

# Abstract

Tumor proliferation is an important biomarker indicative of the prognosis of breast cancer patients. Assessment of tumor proliferation in a clinical setting is a highly subjective and labor-intensive task. Previous efforts to automate tumor proliferation assessment by image analysis only focused on mitosis detection in predefined tumor regions. However, in a real-world scenario, automatic mitosis detection should be performed in whole-slide images (WSIs) and an automatic method should be able to produce a tumor proliferation score given a WSI as input. To address this, we organized the TUmor Proliferation Assessment Challenge 2016 (TUPAC16) on prediction of tumor proliferation scores from WSIs.

The challenge dataset consisted of 500 training and 321 testing breast cancer histopathology WSIs. In order to ensure fair and independent evaluation, only the ground truth for the training dataset was provided to the challenge participants. The first task of the challenge was to predict mitotic scores, i.e., to reproduce the manual method of assessing tumor proliferation by a pathologist. The second task was to predict the gene expression based PAM50 proliferation scores from the WSI.

The best performing automatic method for the first task achieved a quadratic-weighted Cohen's kappa score of $\kappa = 0.567$, 95% CI [0.464, 0.671] between the predicted scores and the ground truth. For the second task, the predictions of the top method had a Spearman's correlation coefficient of $r = 0.617$, 95% CI [0.581 0.651] with the ground truth.

This was the first comparison study that investigated tumor proliferation assessment from WSIs. The achieved results are promising given the difficulty of the tasks and weakly-labeled nature of the ground truth. However, further research is needed to improve the practical utility of image analysis methods for this task.

# 1. Introduction

Tumor proliferation is an important biomarker indicative of the prognosis of breast cancer patients. Patients with high tumor proliferation have worse outcomes compared to patients with low tumor proliferation (van Diest et al., 2004). The assessment of tumor proliferation influences the clinical management of the patient – patients with aggressive tumors are treated with more aggressive therapies and patients with indolent tumor are given more conservative treatments that are preferred because of fewer side-effects (Fitzgibbons et al., 2000).

Tumor proliferation in a clinical setting is traditionally assessed by pathologists. The most common method is to count mitotic figures (dividing cell nuclei) on hematoxylin & eosin (H&E) histological slides under a microscope. The pathologists will assign a mitotic score of 1, 2 or 3, where a score of 3 represents high tumor proliferation. Two other more objective methods that assess tumor proliferation in the breast include immunohistochemical staining for Ki67 protein (Cheang et al., 2009) and the molecular gene expression based PAM50 proliferation score (Heng et al., 2017; Nielsen et al., 2010; The Cancer Genome Atlas Network, 2012). The lack of standardized procedures, debates about clinical utility, issues with Ki67 assay interpretation, and the complex molecular workflow to obtain gene expression has impeded the translation of Ki67 and PAM50 proliferation score for clinical use (Penault-Llorca and Radosevic-Robin, 2017). Ki67 and PAM50 proliferation score are significantly associated with mitotic counts (Heng et al., 2017; Lee et al., 2014), but their agreement is not perfect. There are limited studies investigating the relationship between molecular and morphological data, specifically, none has explored the potential of predicting PAM50 proliferation scores from H&E WSIs.

Although mitosis counting is routinely performed in most pathology practices, this highly subjective and labor-intensive task suffers from reproducibility problems (Veta et al., 2016). One solution is to develop automated computational pathology systems to efficiently, accurately and reliably detect and count mitotic figures on histopathological images. Mitosis detection in WSIs is an active field of research (Albarqouni et al., 2016; Chen et al., 2016, 2016; Li et al., 2018; Tellez et al., 2018, 2018b). This interest was in large part supported by the availability of public datasets in the form of medical image analysis challenges. The first challenge on the topic of on mitosis detection was MITOS 2012 hosted at the International

Conference of Pattern Recognition (ICPR) 2012 (Roux et al., 2013). In 2013, we organized AMIDA13 in conjunction with the International Conference on Medical Image Computing and Computer Assisted Intervention (MICCAI) conference (Veta et al., 2015). Mitosis detection was also one of the tasks of the MITOS-ATYPIA-14 challenge, organized as part of ICPR 2014, with the other task being scoring of nuclear atypia (Roux, 2014).

A large limitation of all previous challenges was that they focused solely on mitosis detection in predetermined tumor regions of interest (ROIs). However, in a real-world scenario, automatic mitosis detection should be performed in WSIs and an automatic method should ideally be able to produce a breast tumor proliferation score given a WSI as input. To address this, we organized the TUmor Proliferation Assessment Challenge 2016 (TUPAC16). The main goal of the challenge was to evaluate (semi-)automatic methods to assess tumor proliferation from WSIs. In this paper, we present an overview of the submitted methods and results of the TUPAC16 challenge.

## 1.1 Challenge format and tasks

The challenge was organized in the context of MICCAI 2016 conference in Athens, Greece. The participants were able to register via the TUPAC16 website[1] six months prior to the MICCAI 2016 conference, allowing ample time to develop their algorithms and submit results. Upon registration, the participants were provided with a training and testing dataset to develop an automatic tumor proliferation scoring method. Two auxiliary datasets that could aid the method development were also provided (see Materials and Methods section). In order to ensure fair and independent evaluation, only the ground truth for the training dataset was provided. The ground truth for the testing dataset was retained by the challenge organizers.

The challenge had two main tasks to predict tumor proliferation. The first task was to predict mitotic scores. In essence, this task aims to reproduce the most common method of assessing tumor proliferation by a pathologist. The second task was to predict the gene expression based PAM50 proliferation scores. While it has been previously shown that PAM50 proliferation scores correlate with manual mitotic scores (Heng et al., 2017), the goal of this task was to determine whether molecular scores can be predicted from tissue morphology/WSIs.

---

[1] http://tupac.tue-image.nl

A third task on mitosis detection was later added to the challenge upon request from the participants. This task was similar and related to the AMIDA13 challenge (Veta et al., 2015). However, due to the auxiliary nature of this task, we will not present an extensive overview in this paper and focus solely on the tumor proliferation assessment from WSIs. In brief, the top scoring method for the third task had an $F_1$ score of 0.652 or mitosis detection. This is a slight improvement over the top scoring method of AMIDA13 challenge which had an $F_1$ score of 0.612. The results table for this task is available on the Results page published on the challenge website.

All participating individuals or teams submitted their results for evaluation on the challenge website. In order to prevent overfitting on the test set, the number of submissions was limited to three per task. All submitted results before the deadline of October 3rd 2016 were presented at the challenge workshop and are included in this paper. Prior to the submission deadline, 159 teams registered on the challenge website. Twelve teams submitted results for the first task[2] and six teams submitted results for the second task.

## 2. Materials and Methods

### 2.1 Main dataset from The Cancer Genome Atlas

The Cancer Genome Atlas (TCGA) Network was established to understand the molecular basis of 33 types of cancer. Specifically, the TCGA breast cancer team utilized genomic, transcriptomic and proteomic profiling technologies to characterize over 1200 invasive breast cancer cases (The Cancer Genome Atlas Network, 2012). Heng *et al*. (2017) subsequently curated a highly detailed histopathological annotation database for a subset of 850 TCGA breast cancer cases and integrated underlying molecular mechanisms with breast cancer morphological features.

---

[2] This overview paper includes he methods of 11 teams. The team with the lowest ranking method for the first task asked to be excluded from the overview paper (the results of this team are visible on the challenge website).

Each case in Heng *et al.* (2017) was represented by one WSI scanned at 40× magnification with an Aperio ScanScore scanner (Gutman et al., 2013). Cases were randomly assigned to a team of 15 international breast pathology experts to assess 12 breast cancer morphological features. Most features adhered to criteria established in clinical practice (Lester et al., 2009), criteria for certain features had to be modified to assess WSIs. In particular, the pathologists developed mitotic count thresholds specific to the TCGA study (Heng et al., 2017) where a score of 1 represents 0 to 5 mitotic counts per 10 HPFS at 40× magnification; a score of 2 represents 6 to 10 mitotic counts per 10 HPFS; and a score of 3 for >10 mitotic counts per 10 HPFS. To perform mitotic counting, the pathologists maximized their window screen size, pulled the zoom bar to the maximum and scanned each WSI to find an area with highest mitotic activity. Each on-screen area at maximum magnification was defined as a high powered field (HPF). The pathologists counted the number of mitoses per field in 10 consecutive fields, excluded fields with scant numbers of tumor cells or necrosis, and the sum over the 10 fields was used to determine the score. Mitotic scores were available for 821 cases, of which 311 were scored by at least two pathologists with an inter-rater reliability Krippendorff's alpha of 0.488 and 78% agreement (Heng et al., 2017). In the 311 cases scored by more than one pathologist, a consensus was formed by taking the most common mitotic score (in case of a tie, the highest mitotic score was taken as the consensus). Gene expression based PAM50 proliferation score was available for all cases. The PAM50 proliferation score is the average expression of 11 proliferation-associated genes part of the PAM50 gene signature: *BIRC5, CCNB1, CDC20, CEP55, MKI67, NDC80, NUF2, PTTG1, RRM2, TYMS* and *UBE2C* (Heng et al., 2017; Nielsen et al., 2010).

Therefore, the main challenge dataset consisted of 821 TCGA WSIs with two types of tumor proliferation data: mitotic score (manual mitosis counting by the pathologists) and PAM50 proliferation score (derived from gene expression). These 821 cases were randomly split into a training ($n = 500$) and testing ($n = 321$) set. In total, there are 383 cases with a mitotic score of 1, 194 cases with score 2 and 244 cases with score 3. The mean PAM50 proliferation score is -0.176 with a standard deviation of 0.428.

**Table 1** – Distribution of the mitotic score (task 1) and PAM50 score (task 2) in the training and testing datasets.

|  | Score 1 | Score 2 | Score 3 | PAM50 score (mean ± STD) |
|---|---|---|---|---|
| Training | 236 (47%) | 117 (23%) | 147 (30%) | -0.166 ± 0.446 |
| Testing | 147 (46%) | 77 (24%) | 97 (30%) | -0.192 ± 0.400 |

**2.2 Auxiliary datasets**

In addition to the main challenge dataset, two auxiliary datasets (ROI and mitosis detection) were also provided to the participants. These two datasets were to facilitate the design of a WSI tumor proliferation scoring system, e.g., by following a two-step approach to emulate how a pathologist would assess a slide for tumor proliferation: identify ROIs followed by mitotic counting.

**ROI dataset**: The ROI auxiliary dataset contained 148 cases which were randomly selected from the training dataset (this auxiliary dataset did not contain WSIs from the test set). A blinded pathology resident annotated three ROIs to indicate where a pathologist might perform mitosis counting, adhering to standard clinical guidelines (Lester et al., 2009). Mitosis counting is performed in tumor regions that have high cellularity and are preferably located at the periphery. Note that these ROIs identified by the pathology resident may not necessarily overlap with the HPFS used by the team of pathologists who graded the mitotic scores in Heng *et al.* (2017). Examples of ROI annotations by the pathology resident in the auxiliary ROI dataset are given in Figure 1.

**Mitosis detection dataset**: The mitosis detection dataset consisted of WSIs from 73 breast cancer cases from three pathology centers with annotated mitotic figures by consensus of three observers. Of the 73 cases, 23 were previously released as part of the AMIDA13 challenge (Veta et al., 2015). These cases were collected from the Department of Pathology at the University Medical Center in Utrecht, The Netherlands. Each case was represented with varying number of HPFS extracted from WSIs acquired with the Aperio ScanScope XT scanner at 40× magnification with a spatial resolution of 0.25 μm/pixel.

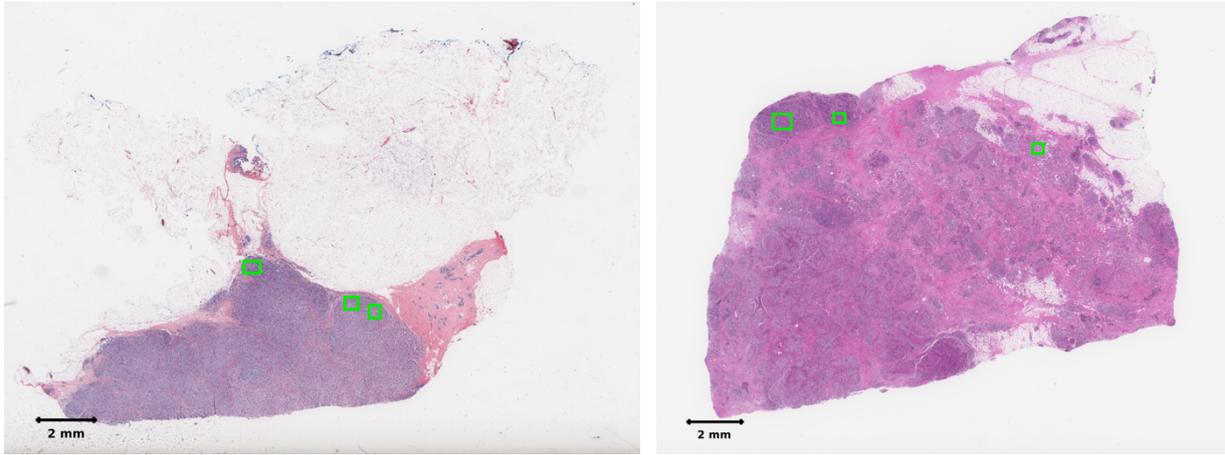

**Figure 1** – Examples of two low magnification whole slide images in the auxiliary region of interest (ROI) dataset annotated with three ROIs (green rectangle boxes) each by a pathology resident. These ROIs represent areas where a pathologist might perform mitosis counting.

The remaining 50 cases previously used to assess the inter-observer agreement for mitosis counting were from two other pathology centers in The Netherlands (Symbiant Pathology Expert Center, Alkmaar and Symbiant Pathology Expert Center, Zaandam) (Veta et al., 2016). Each case was represented by one WSI region with an area of 2 mm². These WSIs were obtained using the Leica SCN400 scanner (40× magnification and spatial resolution of 0.25 μm/pixel). The annotated mitotic figures are the consensus of at least two pathologists, similar to the AMIDA13 challenge. In total, the mitosis detection auxiliary dataset contained 1552 annotated mitotic figures (Figure 2).

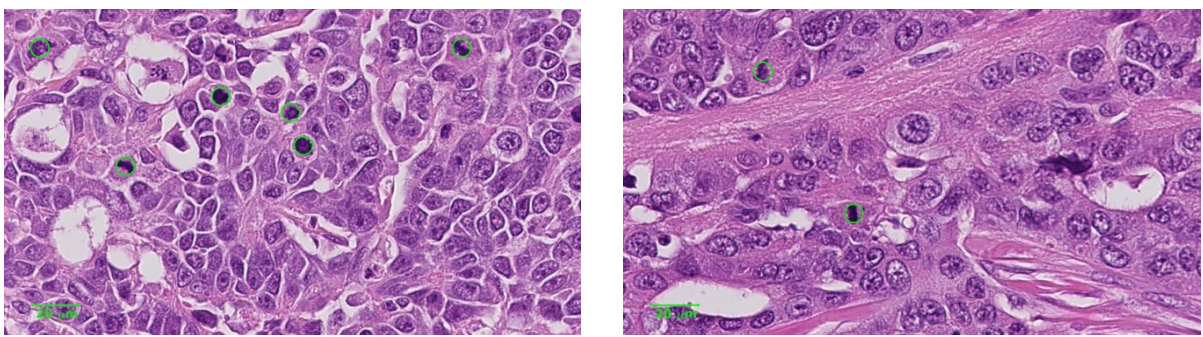

**Figure 2** – Examples from the mitosis detection auxiliary dataset with annotated mitotic figures (green circles). These annotated mitotic figures are the consensus of at least two pathologists.

## 2.3 Summary of the submitted methods

All submitted methods broadly fell into two groups depending on the main strategy to predict proliferation scores from WSIs. The first group of methods followed a pathologist's two-step approach: identify ROIs followed by performing mitosis counting within the selected regions. The prediction for the tumor proliferation scores was based on the response of the mitosis detector. The second group of methods followed a more direct strategy. The first step was also ROI detection, however, mitosis detection was not performed and the prediction of tumor proliferation scores was based on the overall appearance of the ROIs. All teams followed similar strategy for the prediction of both proliferation scores.

With the exception of one team, all participants/teams used deep convolutional neural networks as part of the processing pipeline. Table 2 presents an overview of the main characteristics of all submitted methods in the challenge. The remainder of this subsection summarizes the main characteristics of the proposed methods. A more detailed description of the individual methods can be found in the Supplementary Materials.

**Preprocessing and ROI detection**: One of the major hurdles in histopathology image analysis is the variability of tissue appearance. The staining color and intensity can be significantly different between WSIs due to variation in tissue preparation, staining and digitization processes. To address this, the majority of submitted method performed staining normalization as a preprocessing step. The most commonly used method was the one proposed by Macenko *et al.* (2009). This unsupervised method heuristically estimates the absorbance coefficients for the H&E stains for every image and the staining concentrations for every pixel. Normalization was performed by recomposing the RGB images from the staining concentration maps using common absorbance coefficients.

Since large portions of the WSIs are background, many of the proposed methods resort to tissue segmentation by thresholding such as Otsu's method (Otsu, 1979). The HEIDELBERG team also detected artifacts such as marker annotations or tissue folds based on heuristic mapping that highlights them.

The ROI auxiliary dataset contained non-exhaustive annotations of tumor areas where mitosis detection can be performed. Two teams used this data to train a one-class classifier. The IBM team trained a convolutional autoencoder with the provided ROIs and used the reconstruction error as metric to identify ROIs in test images. BELARUS identified ROIs using the $L_1$ similarity of co-occurrence features to the ground truth ROIs (Kovalev et al., 2001).

SECTRA and WARWICK made in-house annotations of non-ROI regions and used this data to train a supervised ROI detection method (note that the annotated ROIs were used only for training). The method proposed by HARKER also trained a supervised model for ROI detection, however instead of manual annotation for the negative class, HARKER made the assumption that all regions that were not annotated were negative. LUNIT detected ROIs based on cell/nuclei density estimated with CellProfiler (Kamentsky et al., 2011). The MICROSOFT team used manually selected ROIs annotated by an external pathologist[3]. Therefore, their method was classified as semi-automatic.

**Mitosis detection**: All teams that performed mitosis detection as part of the proliferation scoring pipeline used deep convolutional neural networks. Most teams trained a two-class classification model: patches centered at a mitotic figure and background patches. On the testing dataset, the model evaluated every pixel location and produced a mitosis probability map that could be further processed to identify mitotic figures and/or produce a mitotic score for a ROI. The neural network architectures applied to this problem vary from relative "shallow" networks with only a few convolutional layers (CONTEXTVISION and SECTRA) to deep residual neural networks (LUNIT and IBM) (He et al., 2015).

The mitosis detection model by HEIDELBERG did not follow the patch-based approach. They trained a model that incorporated a Hough voting layer – each pixel location predicted the radius and angle to the nearest mitotic figure.

Since mitoses are generally rare events, even in high grade cancers, the mitosis/background classification problem is very unbalanced. In order to remedy this, the majority of submitted methods resorted to two strategies. The first strategy was data augmentation by geometric transformations of the training samples. The mitosis detection problem is invariant to rotations, flipping and small translation and scaling. This can be exploited to create new plausible training samples to enrich the training data.

---

[3] This additional data and the manually annotated ROIs in the testing set used by the MICROSOFT team are available on the TUPAC16 website.

The other strategy was hard negative mining, which was first proposed for mitosis detection by Cireşan *et al.* (2013). With this boosting-like technique, an initial mitosis detection method is trained with random sampling for the background class and then used to detect "difficult" negative instances that are used to train a second method. In practice, models trained with random sampling for the background class result in a large number of false positives since all hyperchromatic objects (e.g. lymphocytes, apoptotic nuclei, necrotic nuclei etc) are detected as mitoses. The output of the initial mitosis detection method can be used to sample such difficult background samples and train a second mitosis detection method. This commonly leads to significant improvements of the mitosis detection accuracy.

**Prediction of tumor proliferation score for Task 1**: CONTEXTVISION, SECTRA, HEIDELBERG and FLORIDA predicted proliferation scores for the first task with heuristic methods based on combining the results from the detection of ROIs and mitoses. For instance, CONTEXTVISION computed a proliferation score for every detected ROI by counting the number of pixels in the mitosis probability map above a certain threshold value that was optimized by cross-validation. A slide-level score was produced by taking the maximum over all ROIs. The final prediction was made by quantizing the slide-level score into one of the three grades based on the grade distribution in the training set. Similarly, SECTRA computed a ROI score that combined the number of detected mitotic figures and the per-pixel average of the mitosis detection model. A slide-level score was computed by averaging the scores for all detected ROIs and then stratified into three categories using two threshold values optimized on the training set. HEIDELBERG computed a slide-level proliferation score as the $95^{th}$-percentile of the mitotic counts for the detected ROIs.

LUNIT, IBM, HARKER and WARWICK predicted the proliferation score with a classifier that used a combination of features based on the output of the mitosis detection method and global ROI features. LUNIT trained a support vector machine (SVM) classification model using a set of features that summarized the statistics of the number of detected mitoses and nuclei in 30 ROIs. IBM trained a random forest classifier using global color and texture features (average intensity of the RGB channels, contrast, energy and homogeneity) and number of detected mitoses at four different detection levels in six ROIs.

Three of the proposed methods (BELARUS, RADBOUD and MICROSOFT) followed a direct strategy for predicting proliferation scores that did not rely on mitosis detection. BELARUS was the only team that did not employ deep neural networks in any part of the processing pipeline and predicted the tumor proliferation score with a linear classifier trained with a set of co-occurrence features (Kovalev et al., 2001). The method submitted by RADBOUD was unique among the submissions since it did not rely on ROI detection. Instead, large image patches from a low magnification level of the WSI (4096×4096 pixels, 5× magnification) were cropped with data augmentation and used as input into a deep neural network model to predict the proliferation score. MICROSOFT computed features in the manually annotated ROIs with a pre-trained ResNet model (He et al., 2015) and then trained a RankSVM (Joachims, 2002) with a linear kernel to make the predictions.

**Prediction of proliferation score for Task 2**: All six teams that participated in the second task used a similar or identical approach as for the first task, e.g. by using a regression instead of a classification model.

## 2.4 Evaluation

The first task was evaluated using the quadratic weighted Cohen's kappa statistic for inter-rater agreement between the ground truth and the predictions. This variant of Cohen's kappa puts higher weight on larger errors in the predicted grade (e.g. "1" instead of "3" or vice versa) that are of higher clinical consequence. The second task was evaluated with the Spearman's correlation coefficient between the prediction and the ground truth PAM50 proliferation scores.

**Table 2** – Summary of the submitted methods.

| Team name | Use of additional training data | Preprocessing | ROI detection | Mitosis detection | Predictions for Task 1 | Predictions for Task 2 |
|---|---|---|---|---|---|---|
| **LUNIT** <br> **Lunit Inc., Korea** | No | Tissue segmentation with Otsu thresholding (Otsu, 1979); staining normalization (Macenko et al., 2009) | Based on cell density estimated with CellProfiler (Kamentsky et al., 2011) | ResNet architecture (He et al., 2015); hard negative mining | SVM classifier with 21 types of features related to cell and mitotic figures density <br><br> Rank for Task 1: 1 | SVM for regression, same features as for Task 1 <br><br> Rank for Task 2: 2 |
| **CONTEXTVISION** <br> **Contextvision, Sweden (SLDESUTO-BOX)** | No | None | Based on heuristic mapping of the color channels that highlights dark tumor areas | Architecture similar to Cireşan *et al.* (2013); hard negative mining | Heuristic based on the response on mitosis detection in the ROIs <br><br> Rank for Task 1: 3 | Same as for Task 1 <br><br> Rank for Task 2: 4 |
| **SECTRA** <br> **Sectra, Sweden** | Yes; non-ROI annotations | None | Based on classification with a four-layer CNN | Six-layer CNN; hard negative mining | Heuristic based on the response on mitosis detection in the ROIs <br><br> Rank for Task 1: 4 | *d.n.p.* |
| **HEIDELBERG** <br> **University of Heidelberg, Germany** | No | Artifact detection based on heuristic mapping of the color channels that highlights ink and tissue folding | Based on heuristic mapping of the color channels that highlights dark tumor areas | Novel architecture that combine residual networks with Hough voting (Wollmann and Rohr, 2017); hard negative mining | Thresholds for the number of detected mitotic figures optimized using the quadratic weighted Cohen's kappa score <br><br> Rank for Task 1: 5 | *d.n.p.* |
| **IBM** <br> **IBM Research Zurich and Brazil** | Yes; ICPR 2012 and 2014 datasets | Staining normalization (Macenko et al., 2009) | One-class classification based on the reconstruction error of convolutional autoencoders | Wide residual network 22-2 architecture (Zagoruyko and Komodakis, 2016); hard negative mining | Random forest classifier using color, texture and number of mitoses features in the detected ROIs <br><br> Rank for Task 1: 6 | *d.n.p.* |

| Team | | Semi-automatic | Pre-processing | ROI detection | Mitosis detection | Task 1 | Task 2 |
|---|---|---|---|---|---|---|---|
| **HARKER** <br><br> The Harker School, United States | | No | Tissue segmentation with Otsu thresholding (Otsu, 1979); staining normalization (Ehteshami Bejnordi et al., 2015) | Based on classification with four neural network architectures: GoogLeNet (Szegedy et al., 2014), ResNet-34 (He et al., 2015), VGG-13 (Simonyan and Zisserman, 2014) and custom architecture | Custom CNN architecture; hard negative mining; end-to-end models for predicting the mitotic score from the ROIs | Combination random forest, SVM and gradient boosting classifiers using a combination of features from the ROI detection, mitosis detection and end-to-end models <br><br> Rank for Task 1: 7 | Regression with the same features as for Task 1 <br><br> Rank for Task 2: 6 |
| **BELARUS** <br><br> Belarus National Academy of Sciences | | No | Staining decomposition; the hematoxylin channel was used in all subsequent processing | One-class classification based on $L_1$ similarity with co-occurrence features (Kovalev et al., 2001) | n/a | Direct prediction using a linear classifier with co-occurrence features (average of prediction for 20 ROIs) <br><br> Rank for Task 1: 8 | Regression with the same features as for Task 1 <br><br> Rank for Task 2: 5 |
| **RADBOUD** <br><br> Radboud UMC Nijmegen, The Netherlands | | No | Tissue detection with thresholding; Staining normalization (Ehteshami Bejnordi et al., 2015) | n/a | n/a | End-to-end prediction using a custom CNN architecture (average of 500 predictions for randomly cropped regions) <br><br> Rank for Task 1: 9 | Same as for Task 1 <br><br> Rank for Task 2: 3 |
| **FLORIDA** <br><br> University of South Florida, United States | | No | Staining normalization (Macenko et al., 2009) | Based on heuristic mapping of the color channels that highlights dark tumor areas | AlexNet architecture (Krizhevsky et al., 2012) | Heuristic thresholds for the number of detected mitotic figures in the ROIs <br><br> Rank for Task 1: 10 | d.n.p. |
| **WARWICK** <br><br> University of Warwick, United Kingdom | | Yes; non-ROI annotations | None | Based on tumor segmentation with U-Net-like architecture (Ronneberger et al., 2015) | Two-stage CNN detector | Random forest classifier using number of mitoses features in the detected ROIs <br><br> Rank for Task 1: 11 | d.n.p. |
| **MICROSOFT** <br><br> Microsoft Research Asia, China | | No | Staining normalization (Macenko et al., 2009) | Manual ROI selection | n/a | RankSVM with linear kernel (Joachims, 2002); feature extraction was done with ResNet (He et al., 2015) and P-norm pooling(Xu et al., 2015) <br><br> Rank for Task 1: 2 (note: semi-automatic method) | Same as for Task 1 <br><br> Rank for Task 2: 1 (note: semi-automatic method) |

*d.n.p.* – did not participate for this task
*n/a* - not applicable for this method

# 3. Results

The results for the first task in the challenge (prediction of tumor proliferation score based on mitosis counting) are summarized in Table 3. The top performing method was by LUNIT with a quadratic weighted kappa statistic of $\kappa = 0.567$, 95% CI [0.464, 0.671]. The semi-automatic method by MICROSOFT and the method submitted by CONTEXTVISION had similar performances of $\kappa = 0.543$, 95% CI [0.422, 0.664] and $\kappa = 0.534$, 95% CI [0.422, 0.646], respectively. Table 4 presents the confusion matrices of the predictions using the methods by LUNIT, MICROSOFT and CONTEXTVISION (the confusion matrices for all methods can be found in the Supplementary Materials), along with an ensembling by average voting of the top 3 automatic methods (see section 3.1). As evident by the per-class accuracies, mitotic score 2 was the most commonly misclassified (per-class accuracy of 17%, 50% and 31% for LUNIT, MICROSOFT and CONTEXTVISION).

**Table 3** – Results for Task 1.

|    | Team          | $\kappa$ † | 95% CI         |
|----|---------------|------------|----------------|
| 1  | LUNIT         | 0.567      | [0.464, 0.671] |
| 2  | MICROSOFT*    | 0.543      | [0.422, 0.664] |
| 3  | CONTEXTVISION | 0.534      | [0.422, 0.646] |
| 4  | SECTRA        | 0.462      | [0.340, 0.584] |
| 5  | HEIDELBER     | 0.417      | [0.293, 0.540] |
| 6  | IBM           | 0.385      | [0.266, 0.504] |
| 7  | HARKER        | 0.367      | [0.242, 0.492] |
| 8  | BELARUS       | 0.321      | [0.190, 0.452] |
| 9  | RADBOUD       | 0.290      | [0.171, 0.409] |
| 10 | FLORIDA       | 0.177      | [0.052, 0.302] |
| 11 | WARWICK       | 0.159      | [0.023, 0.294] |

\* Semi-automatic method
† Quadratic weighted Cohen's kappa statistic

Table 5 summarizes the results of PAM50 proliferation score prediction. The best performance was achieved by the semi-automatic method by MICROSOFT ($r = 0.710$, 95% CI [0.681 0.737]). The best scoring automatic method was LUNIT with a Spearman correlation coefficient between the ground truth and predicted scores of $r = 0.617$, 95% CI [0.581 0.651], followed by RADBOUD with $r = 0.516$, 95% CI [0.474 0.556]. The scatterplots between the ground truth and predicted PAM50 proliferation scores for the MICROSOFT, LUNIT and RADBOUD methods are in Figure 3. Scatterplots for all methods can be found in the Supplementary Materials along with evaluation of the agreement between the prediction and the ground truth in terms of the intraclass correlation coefficient.

**Table 4** –Confusion matrices for the LUNIT (A), MICROSOFT (B) and CONTEXTVISION (C) methods and the ensembling by average voting of the top three automatic methods (LUNIT, CONTEXTVISION and SECTRA; D).

**A** LUNIT

|  |  | Predicted 1 | Predicted 2 | Predicted 3 | Acc. |
|---|---|---|---|---|---|
| Ground truth | 1 | 117 | 11 | 19 | 80% |
|  | 2 | 40 | 13 | 24 | 17% |
|  | 3 | 16 | 8 | 73 | 75% |

$\kappa = 0.567$
95% *CI* [0.464, 0.671]

**B** MICROSOFT*

|  |  | Predicted 1 | Predicted 2 | Predicted 3 | Acc. |
|---|---|---|---|---|---|
| Ground truth | 1 | 99 | 44 | 4 | 67% |
|  | 2 | 26 | 39 | 12 | 50% |
|  | 3 | 11 | 46 | 40 | 41% |

$\kappa = 0.543$
95% CI [0.422, 0.664]

**C** CONTEXTVISION

|  |  | Predicted 1 | Predicted 2 | Predicted 3 | Acc. |
|---|---|---|---|---|---|
| Ground truth | 1 | 103 | 29 | 15 | 70% |
|  | 2 | 35 | 24 | 18 | 31% |
|  | 3 | 14 | 22 | 61 | 62% |

$\kappa = 0.534$
95% CI [0.422, 0.646]

**D** Ensembling by average voting of LUNIT, CONTEXTVISION and SECTRA

|  |  | Predicted 1 | Predicted 2 | Predicted 3 | Acc. |
|---|---|---|---|---|---|
| Ground truth | 1 | 107 | 36 | 4 | 73% |
|  | 2 | 36 | 29 | 12 | 38% |
|  | 3 | 7 | 40 | 50 | 52% |

$\kappa = 0.613$
95% CI [0.504, 0.722]

* Semi-automatic method
Acc. – per-class accuracy for the three mitotic scores

**Table 5** – Results for Task 2.

|  | Team | $r^{\dagger}$ | 95% CI |
|---|---|---|---|
| 1 | **MICROSOFT*** | 0.710 | [0.681 0.737] |
| 2 | **LUNIT** | 0.617 | [0.581 0.651] |
| 3 | **RADBOUD** | 0.516 | [0.474 0.556] |
| 4 | **CONTEXTVISION** | 0.503 | [0.460 0.544] |
| 5 | **BELARUS** | 0.494 | [0.451 0.535] |
| 6 | **HARKER** | 0.474 | [0.429 0.516] |

* Semi-automatic method
† Spearman correlation coefficient

## 3.1 Method ensembling

Exploratory experiments with model ensembling were performed by averaging the results of the top three automatic methods (the semi-automatic methods by MICROSOFT were excluded from this analysis). For the first task, the predicted scores from the top three automatic methods (LUNIT, CONTEXTVISION and SECTRA) were averaged and rounded to the nearest integer. This resulted in a score of $\kappa = 0.613$, 95% CI [0.504, 0.722], which is an improvement by 0.046 over the best individual method by LUNIT ($\kappa = 0.567$, 95% CI [0.464, 0.671]). For the second task, the predictions by the top three automatic methods (LUNIT, RADBOUD and CONTEXTVISION) were first scaled to zero-mean and unit-variance and then averaged. This was necessary in order to account for the different scales of the predictions. The combined prediction resulted in a Spearman correlation coefficient of $r = 0.682$, 95% CI [0.651 0.711], which is an improvement by 0.065 over the best individual method by LUNIT ($r = 0.617$, 95% CI [0.581 0.651]; Figure 3D).

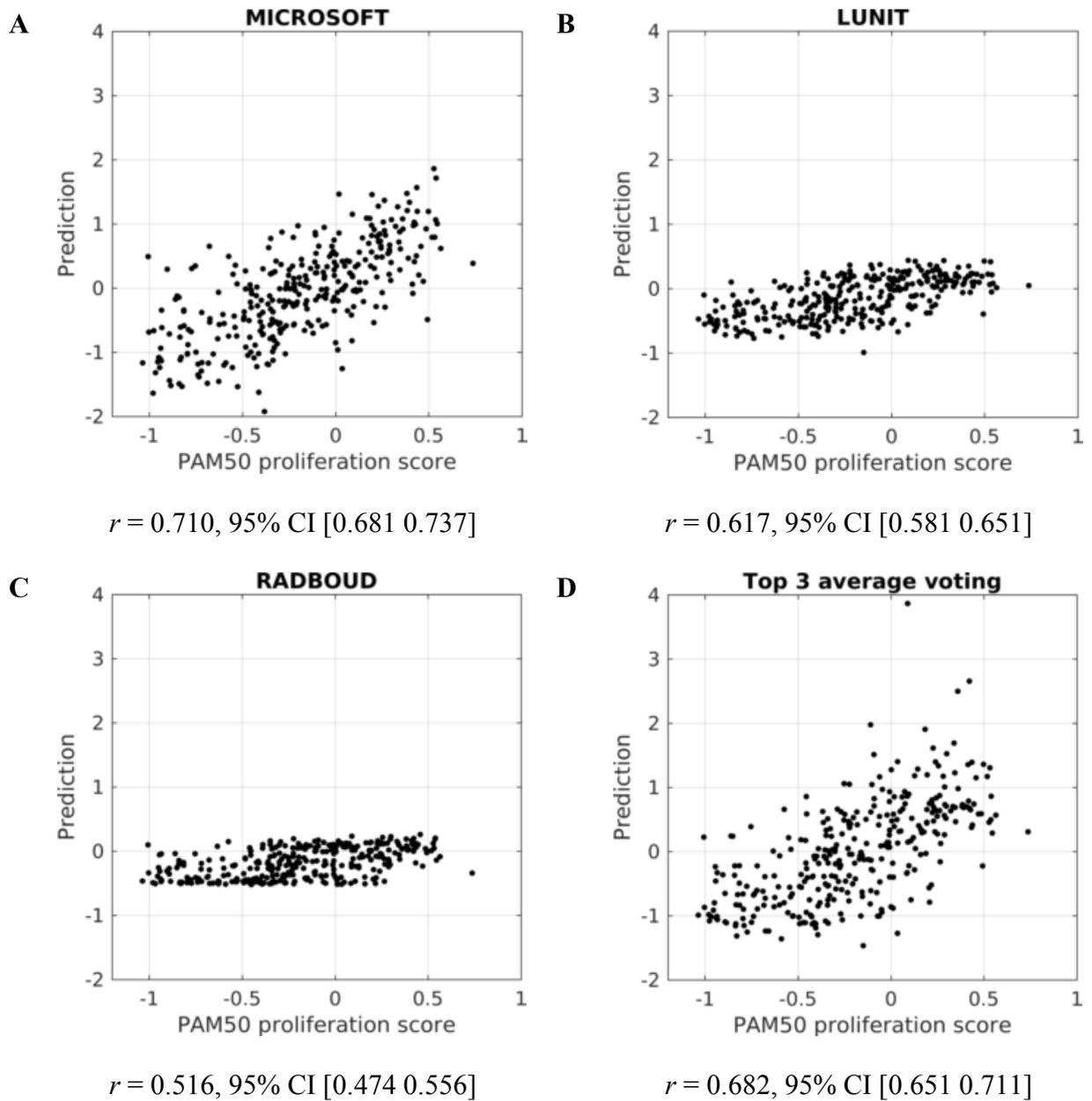

**Figure 3** − Results for Task 2. Scatter plots for the MICROSOFT (A), LUNIT (B) and RADBOUD (C) methods and the average voting of the top three automatic methods (LUNIT, RADBOUD and CONTEXTVISION; D).

## 4. Discussion

Tumor proliferation is an important pathological assessment that aids the clinical management of cancer patients. The current method to assess tumor proliferation is manual mitosis counting by pathologists. This process is highly subjective and time consuming. An automatic computational pathology proliferation assessment method will save time and lead to the standardization of mitotic scores across institutions. The TUPAC16 challenge was created to advance the state of the art in automatic assessment of tumor proliferation from WSIs and improve upon previous challenges that focused solely on mitosis detection.

The first task of the challenge was to predict a mitosis-based tumor proliferation score. The best performing method (LUNIT) achieved a quadratic-weighted Cohen's kappa score of $\kappa = 0.567$, which signifies a moderate agreement with the manual ground truth. This $\kappa$ score was lower compared to previous work. In Veta *et al.* (2016), the inter-observer agreement between pathologists was estimated between $\kappa = 0.792$ and $\kappa = 0.893$. The higher $\kappa$ agreement in Veta *et al.* was most likely due to the three pathologists performing mitosis counting in a predefined area, which considerably increases the chances for a concordant score as it eliminates the tumor heterogeneity factor, compared to the first task of the challenge where the teams may have predicted mitotic scores in ROIs different from the pathologists in Heng *et al.* (2017). All three best performing methods for the first task also made a substantial number of errors whereby the predicted and ground truth scores differed by two (Table 4). Such discordance may lead to more severe clinical implications, however similar errors can also potentially occur with manual scoring (Al-Janabi et al., 2013; Robbins et al., 1995), although to a lesser extent.

The manual scoring of tumor proliferation by mitosis counting involves a multi-scale analysis of the tissue. Thus, training an automatic method that predicts mitotic scores using only global, slide-level annotations is a challenging task. It should also be noted that the mitosis detection auxiliary dataset, which was used by the majority of teams to train a mitosis detector that formed the basis of the proliferation scoring models, was obtained from three Dutch medical centers and are different from the TCGA USA medical institutions that provided the main dataset. This constitutes a domain shift when the mitosis detector trained with the auxiliary dataset is applied to main dataset, which further increases the difficulty of this task.

The top performing automatic methods in the first task followed a two-stage approach that emulates the scoring by pathologists. However, the individual building blocks vary between the methods. While the method by LUNIT performs mitosis detection with a very deep ResNet architecture (He et al., 2015), the methods by SECTRA and CONTEXTVISION that achieved a comparable performance used a comparatively "shallower" neural network architecture. LUNIT use a staining normalization approach (Macenko et al., 2009) to standardize the appearance of the tissue prior to further processing and the mitosis detector used by SECTRA works on grayscale images at a two times reduced resolution (0.5 µm/pixel). The top three automatic methods for the first task used different ROI detection methods (heuristic color channel mapping, cell-density based detection and CNN classifier) and different methods for computing a slide-level proliferation score (SVM classifier, heuristic based on the response of the mitosis detector). The varied method design is likely responsible for the performance boost when ensembling the predictions of the top three methods ($\kappa$ = 0.613 for the average voting).

The second task of the challenge has the built-in hypothesis that the molecular PAM50 proliferation score can be predicted from WSIs. The MICROSOFT, NIJMEGEN and BELARUS methods predicted the tumor proliferation scores for both tasks using region-level features, without resorting to mitosis detection as an intermediate step. Although their methods worked particularly well for the second task and achieved good correlation, the best performing automatic method by LUNIT still relied on mitosis counting. The results from the second task accepted our hypothesis that the molecular PAM50 proliferation score could be predicted from WSIs. Predicting molecular scores from WSIs could potentially be a new clinical and research tool to assess tumor proliferation.

All proposed methods except one can be characterized as automatic as they do not require manual input at test time. The MICROSOFT method is semi-automatic as it requires the ROI regions to be manually selected at test-time. While such an approach has more limited utility compared to automatic methods, it can still be valuable in clinical practice. For example, pathologists can request the method to be executed on the same ROIs where they assessed mitotic counts.

## 4.1 Recommendations for future work and conclusion

TUPAC16 was the first challenge to predict tumor proliferation scores from WSIs. The main goal was to gain insights into automatic solutions for this problem and set the state of the art. As tumor proliferation is an important prognostic biomarker for breast cancer, we expect that this topic will continue to be of relevance in the future. In this subsection, we provide recommendations for subsequent research and challenges on this topic.

**Modular submission format**: The majority of the submitted methods used a pipeline that consisted of four major processing steps: pre-processing, ROI detection, mitosis detection and slide-level prediction. Due to the heterogeneity of the top scoring methods, it is difficult to draw conclusions about design choices that positively impacted the performance of the methods. One of the major drawbacks of the TUPAC16 challenge setup was that it was not feasible to conduct quantitative evaluation on how each image analysis pipeline component impacted the performances of the methods. One solution for future challenges is to request submissions in a modular Docker format. With this setup, the different processing blocks will be submitted as separate Docker containers that are combined to produce the final submission. The use of Docker containers for submission can improve the reproducibility of the submissions and reduce the chance of cheating (Maier-Hein et al., 2018). The modular format can enable evaluating methods that are combinations of building blocks submitted by different teams and facilitate marginalization of the impact of individual design choices, e.g., by evaluating all methods on a predefined set of ROIs or with standardized/baseline mitosis detection.

**Evaluation on datasets from external domains:** Systematic differences in the appearance of the WSIs obtained from different pathology laboratories is one of the major hurdles in histopathology image analysis. Future challenges and research on this topic should have an experimental setup to evaluate the performance of the methods under a domain shift. Minimally, the independent testing set should include a subset of cases from a domain that was not included in the training set (e.g. a different pathology laboratory or scanner manufacturer).

**Investigate the relationship between global/regional features and proliferation:** A particularly interesting finding of this challenge was that both proliferation scores can be predicted with reasonable accuracy from ROI-level features, without resorting to mitosis detection. Future research efforts should focus on investigating the relationship between global or regional image features and tumor proliferation, such as visualizing the learned features that are predictive of high tumor proliferation. This can be particularly of interest for the PAM50 proliferation score as it can establish a relationship between molecular and morphological tissue characteristics.

**Evaluation in terms of prognostication:** The ultimate goal of tumor proliferation assessment is to guide clinical management and predict patient outcome. We recommend that future work on this topic also evaluate the proposed methods in terms of predicting the overall or disease-specific survival of breast cancer patients.

The performance of the automatic and semi-automatic methods submitted to this challenge did not reach a level that is sufficient to be used as a "second opinion" score. The results from the challenge are promising given the difficulty of the tasks and weakly-labeled nature of the ground truth, and have provided valuable insight into this problem. However, further research is needed to improve the practical utility of image analysis methods.

# Supplementary Material

**S1** Method description, available at:

http://tupac.tue-image.nl/system/files/Supplementary%20material%201%20%28rev%29.pdf

**S2** Results for all submitted methods, available at:

http://tupac.tue-image.nl/system/files/Supplementary%20material%202%20%28rev%29.pdf